\def\year{2019}\relax
\documentclass[letterpaper]{article} 
\usepackage{aaai19}  
\usepackage{times}  
\usepackage{helvet}  
\usepackage{courier}  
\usepackage{url}  
\usepackage{graphicx}  
\usepackage{amssymb}
\usepackage{amsmath}
\usepackage{minipage}
\usepackage{booktabs}
\usepackage{amsthm}
\usepackage{multirow}

\theoremstyle{definition}
\newtheorem{example}{Example}

\newcommand{\ptitle}[0]{Generalized Ternary Connect: End-to-End Learning and Compression of Multiplication-Free Deep Neural Networks}
\newcommand{\pauthor}[0]{Samyak Parajuli \and Aswin Raghavan \and Sek Chai\\
firstname.lastname@sri.com\\
SRI International, Princeton, NJ, USA}

\begin{document}
\title{\ptitle{}}
\author{\pauthor{}}
\maketitle

\begin{abstract}
The use of deep neural networks in edge computing devices hinges on the balance between accuracy and complexity of computations. 
Ternary Connect (TC) \cite{lin2015neural} addresses this issue by restricting the parameters to three levels $-1, 0$, and $+1$, thus eliminating multiplications in the forward pass of the network during prediction. We propose Generalized Ternary Connect (GTC), which allows an arbitrary number of levels while at the same time eliminating multiplications by restricting the parameters to integer powers of two. The primary contribution is that GTC learns the  number of levels and their values for each layer, jointly with the weights of the network in an end-to-end fashion. 
Experiments on MNIST and CIFAR-10 show that GTC naturally converges to an ``almost binary'' network for deep classification networks (e.g. VGG-16) and deep variational auto-encoders, with negligible loss of classification accuracy and comparable visual quality of generated samples respectively. We demonstrate superior compression and similar accuracy of GTC in comparison to several state-of-the-art methods for neural network compression. We conclude with simulations showing the potential benefits of GTC in hardware.
\end{abstract}

\section{Motivation and Prior Work}
\label{sec:intro}

The demonstrated success of deep learning in vision, speech and planning domains can be leveraged in practical applications involving edge computing devices only if the requirements of high power, large memory and high-end computational resources are significantly reduced. A common solution that circumvents all three of these issues is to replace multiplications in neural networks with a cheap operation such as bit shifts \cite{marchesi1993fast,kwan1993multiplierless,courbariaux2015binaryconnect,lin2015neural,miyashita2016convolutional} or integer arithmetic \cite{raghavan2017bitnet,wu2018training}, the reason being the ever-increasing number of parameters in deep neural networks (projected to reach 1 trillion parameters within the next decade) and bulky power-hogging digital multipliers. Convolutional networks are especially expensive since the number of multiplications grows with the number of filters and the size of filters, in addition to the size of the input. 

In this paper, we focus on replacing multiplications with logical bit shift operations, which can be implemented efficiently in hardware using shift registers. While prior work \cite{marchesi1993fast,kwan1993multiplierless,courbariaux2015binaryconnect,lin2015neural,miyashita2016convolutional} has noted this possibility, all of the above are restricted to a fixed range of shifts. For example, TC \cite{lin2015neural} restricts all parameters to $0, \pm 1$ (ignoring the scaling factor), which corresponds to the set of operations of shifting by zero and inverting a sign bit, whereas \cite{kwan1993multiplierless} uses a maximum shift of four, and \cite{miyashita2016convolutional} uses a maximum of five. A bit more generally, the majority of prior work on neural network compression e.g.\! \cite{han2015deep,polino2018model,achterhold2018variational} choose the numbers of bits of precision manually. 

The proposed algorithm, Generalized Ternary Connect (GTC), learns simultaneously the number of bits and the values of the quantized weights. The weights are constrained to integer powers of two, which reduces floating point multiplication by the weights to bit shifts. We view the range of bit shifts as the number of bits of precision. Our second contribution is a novel parametrization of such a set of weights with trainable parameters that allow the quantization function to take a variety of shapes, ranging from a hyperbola, to a Heaviside function, to a hyperbolic tangent function or simply an identity mapping. These additional, trainable parameters significantly generalize TC and prior work on learning neural networks with weights that are powers of two. To the best of our knowledge, this is a novel end-to-end algorithm that jointly optimizes learning and compression from the standpoint of increased performance after training.

We evaluate GTC on two tasks: supervised image classification and unsupervised image generation with a variational auto-encoder (VAE) \cite{kingma2013auto}. We show preliminary results on the MNIST and CIFAR-10 image datasets. While prior work on low precision deep neural networks has focused on classification problems, these experiments extend the demonstration to generative models, which can lead to different insights for future work. In addition, we demonstrate a novel variant of GTC that can be used as a  generic optimizer for training neural networks, often outperforming Stochastic Gradient Descent (SGD). On the classification tasks, we compare GTC to several state-of-the-art neural network compression and low-precision approaches in Section \ref{sec:prior}. Among the compared methods, GTC shows the best compression with minimal drop in accuracy while, unlike most of the prior work, producing a neural network whose forward pass does not require any multiplications. 

\section{Formulation}

We use boldface uppercase symbols to denote tensors and boldface lowercase symbols to denote vectors. Let $\{\mathbf{W}^{(1)}, \ldots, \mathbf{W}^{(N)}\}$ be the set of parameters of a deep neural network with $N$ layers. Let $\tilde{\mathbf{W}}^{(l)}$ be the quantized tensor corresponding to $\mathbf{W}^{(l)}$. The output of a deep neural network can be specified as $X^{(l+1)} = f(X^{(l)}; W^{(l)})$ for $l=1,2,\ldots,N$, where $X^{(l)}$ is the input to the $l^{\text{th}}$ layer and $X^{(l+1)}$ is the output (aka activation) of the $l^{\text{th}}$ layer (and the input to the $l+1^{\text{th}}$ layer). Deep neural networks primarily consist of the following layers.
\begin{flalign*}
\mathbf{X}^{(l+1)} &= W^{(l)} . \mathbf{X}^{(l)} \text{(Fully Connected)}\\
\mathbf{X}^{(l+1)} &= W^{(l)} \circledast \mathbf{X}^{(l)} \text{(Convolution)}\\ 
\mathbf{X}^{(l+1)} &= \frac{\gamma^{(l)}}{\sigma^{(l)}} (\mathbf{X}^{(l)}-\mu^{(l)})+\beta^{(l)} \text{(Batch Normalization)}
\end{flalign*}
It is clear that (scalar) multiplications constitute the majority of the computational cost of the forward pass during inference and prediction. For brevity, we omit the bias parameters throughout the paper. In our experiments, GTC outputs weights and biases in the form of bit shifts. 
\subsection{Quantization} 
Ternary Connect (TC) \cite{lin2015neural} replaces multiplications with logical bit shift operations by quantizing weights to the three values $0$ and $\pm 1$ (up to constant). Our goal is to generalize TC by restricting weights to the feasible set
\begin{equation}
\mathcal{S} = \begin{array}{cc}
\{0\} \cup \{\pm 2^k &: k \in \mathbb{Z}\}
\end{array},
\label{eq:feas_set}
\end{equation}
where $\mathbb{Z}$ is the set of integers. In other words, the weights can be $0, \pm 1, \pm 2, \pm 4, \ldots$ and $\pm 1/2, \pm 1/4, \pm 1/8, \ldots$. The generalization to powers of two allows floating-point multiplications to be implemented as logical bit shifts. For a floating point number $x = m.2^e$ with mantissa $m$ and exponent $e$, its product with a weight $w = 2^k \in \mathcal{S}$ is given by
\begin{align}
w \times x &= 2^k \times x = (x << k)\\
m.2^e \times 2^k &= m.2^{e + k} \label{eq:mult_shift_add}
\end{align}
Apparent from the above, in this paper we do not alter the representation of the activation $x \in \mathbf{X}$, although the same techniques developed herein can be extended. Similarly, we do not consider the multiplications in the backward pass during gradient-based training, however a technique similar to Quantized Back-Propagation \cite{lin2015neural} can be developed.
Let $\tilde{w}=\pm 2^k$ denote the quantized weight corresponding to some $w \in \mathbf{W}$. The starting point for our quantization scheme is the identity transformation
\begin{equation}
\tilde{w} = sign(w) 2^{\log_2{|w|}}.
\label{eq:identity_transform}
\end{equation}
where the $sign$ operator returns the sign of the operand, or returns zero if the operand is close to zero. Empirically, we found that the leading sign of $\tilde{w}$ needs to match the sign of $w$ for good performance.

The constraint that $k \in \mathbb{Z}$ is usually enforced through rounding or stochastic rounding \cite{courbariaux2015binaryconnect}. In our case, either rounding operator can be applied to the exponent $\log_2{|w|}$. We introduce novel parameters $\mathbf{\theta}^{(l)}$ that are specific for each layer, and formulate the quantization function as
\begin{equation}
\tilde{w} = sign(w) 2^{round(q(w;\mathbf{\theta}^{(l)}))}.
\end{equation}
Here $q(w, \theta)$ can be any transformation of the parameters $\mathbf{W}$. Whereas prior work has predominantly used fixed quantization schemes, future work may explore schemes where the quantizer can potentially be an entire neural network. Such static quantization schemes are used because they asymptotically  minimize the error introduced due to quantization of the parameters \cite{gish1968asymptotically}, and many previous works use quantization error as an objective criterion \cite{han2015deep,raghavan2017bitnet}. However, when the numbers and values of quantization levels are learned, the effect of quantization error on the accuracy is not clear. Furthermore, the benefits due to reduced computational and memory overhead may be worth a small loss in accuracy due to ignoring quantization error. In this paper, we explore a linear function as the choice of $q$. 

\begin{align}
q(w;\theta) &= \theta_1 + \theta_2\log_2|w| \label{eq:quant_exponent} \\
\tilde{w} &= sign(w) 2^{round(\theta_1+\theta_2\log_2{|w|})} \label{eq:quant}
\end{align}

\begin{figure}
    \centering
    \includegraphics[width=0.8\linewidth, keepaspectratio]{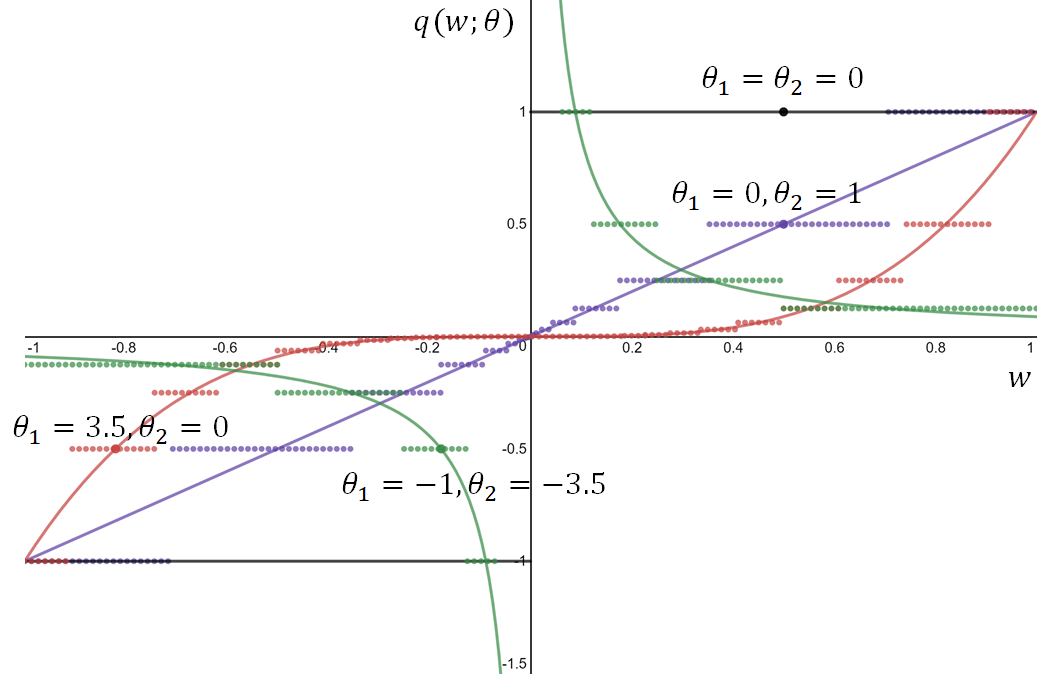}
    \caption{Plot showing different instantiations of $q(w;\theta)$ for different values of $\theta$. Dotted lines shows the effect of rounding as in $round(q(w;\theta))$.}
    \label{fig:quant_example}
\end{figure}

Despite the simple linear form, these $\theta$ parameters allow a very powerful family of transformations. Figure \ref{fig:quant_example} illustrates four example instantiations of $q(w;\theta)$ for some $w \in [-1,1]$. The trivial case is when $\theta_1=0$ and $\theta_2=1$ giving the identity transform. The case when $\theta_1=\theta_2=0$ corresponds to ternary quantization based on the sign of the weight $w$. The case when $\theta_1 < 0, \theta_2 < 0$ gives a family of hyperbolas whose foci and eccentricity are controlled by $\mathbf{\theta}$. The case when $\theta_1 > 0$ gives a family of hyperbolic tangent functions. To the best of our knowledge, this is a novel  parameterization of the set $\mathcal{S}$ (\ref{eq:feas_set}). These additional trainable parameters significantly generalize TC and prior work on learning neural networks with weights that are powers of two \cite{miyashita2016convolutional,marchesi1993fast,kwan1993multiplierless,lin2015neural,courbariaux2015binaryconnect}. Next we describe a novel notion of the bit precision stemming from (\ref{eq:quant}).
\begin{example}
Suppose $\mathbf{W}$ be the matrix, 
$\left(\begin{array}{cccc}
   2.5  & 1 & 1.3 & 0.75 \\
    1 & -2.5 & -1.2 & -0.9
\end{array}\right)$
with $\theta_1=-1$ and $\theta_2=-3.5$, we get $\tilde{W}$ as 
$\left(\begin{array}{cccc}
   2^{-6}  & 2^{-1} & 2^{-2} & 2^0\\
   2^{-1}  & -2^{-6} & -2^{-2} & -2^0
\end{array}\right)$.
Note that the value of $2.5$ was quantized to $2^{-6}=0.015625$ with large quantization error. In our experiments we find that the quantization error is not very important for high accuracy, and that the learned values of $\theta$ are non-trivial. 
\end{example}

\subsection{Compression}
\label{sec:compression}

Consider the compression of a deep neural network whose parameters are of the form of (\ref{eq:quant}) using a fixed length code. Apart from the leading sign bit, each exponent can be stored as an integer. The maximum value of the exponent is $127$, so one might use an $8$-bit integer for a 4X model compression compared to the storage of floating point parameters. Typically, parameters of deep networks should not be allowed to reach such large values as $2^{127}$. Thus, the natural source of compression is the range of values taken by the exponent over all the parameters of the network. Alternatively, one might use the cardinality of the set of values taken by the exponent, but we adopt the former due to its simplicity and differentiability. For each layer $l$, we define the number of bits as shown in (\ref{eq:quant_bits}).
\begin{align}
    m &= \min_{w \in \mathbf{W}^{(l)}} round(q(w; \theta)) \label{eq:quant_min}\\
    M &= \max_{w \in \mathbf{W}^{(l)}} round(q(w; \theta)) \label{eq:quant_max}\\
    \text{bits}^{(l)} &= 1+\lceil \log_2{(M-m+1)} \rceil \label{eq:quant_bits}
\end{align}
The leading term of one accounts for the sign bit, whereas $M-m$ accounts for the range of all the exponents, sufficient to represent all exponents between the maximum and minimum exponent. In some cases, this range may not include zero, whereas $\tilde{w}$ may be zero due to the sign function in (\ref{eq:quant}). Therefore, we add a one and take the logarithm (base 2) to finally arrive at the number of bits. 
\begin{example}
Continuing with our previous example, suppose that $\tilde{W}$ consists of $\{\pm 0.015625, 0.5, \pm 0.25, \pm 1\}$$=\{\pm 2^{-6}, 2^{-1}, \pm 2^{-2}, \pm 2^0\}$, we get $\text{bits}=1+\lceil \log_2{(0-(-6)+1)} \rceil=1+\log_27=4$ bits. Note that this is not an optimal code e.g. we could represent this set of exponents with $2$ bits (excluding the sign bit).
\end{example}
\section{Joint Learning and Compression}
\label{sec:learning}
The difficulty in directly training GTC stems from the discontinuities of the rounding function in (\ref{eq:quant}), (\ref{eq:quant_min}), (\ref{eq:quant_max}) and the ceiling function in (\ref{eq:quant_bits}). One approach to deal with discrete neurons is to use Straight Through Estimation (STE) \cite{bengio2013estimating}. In Section \ref{sec:experiments}, we show the performance of a variant of GTC trained using the biased straight-through estimator, which presumes the gradients of the discrete functions are identity functions. We use this as a baseline for the following training methodology. 

\begin{figure}
    \centering
    \includegraphics[width=\linewidth, keepaspectratio]{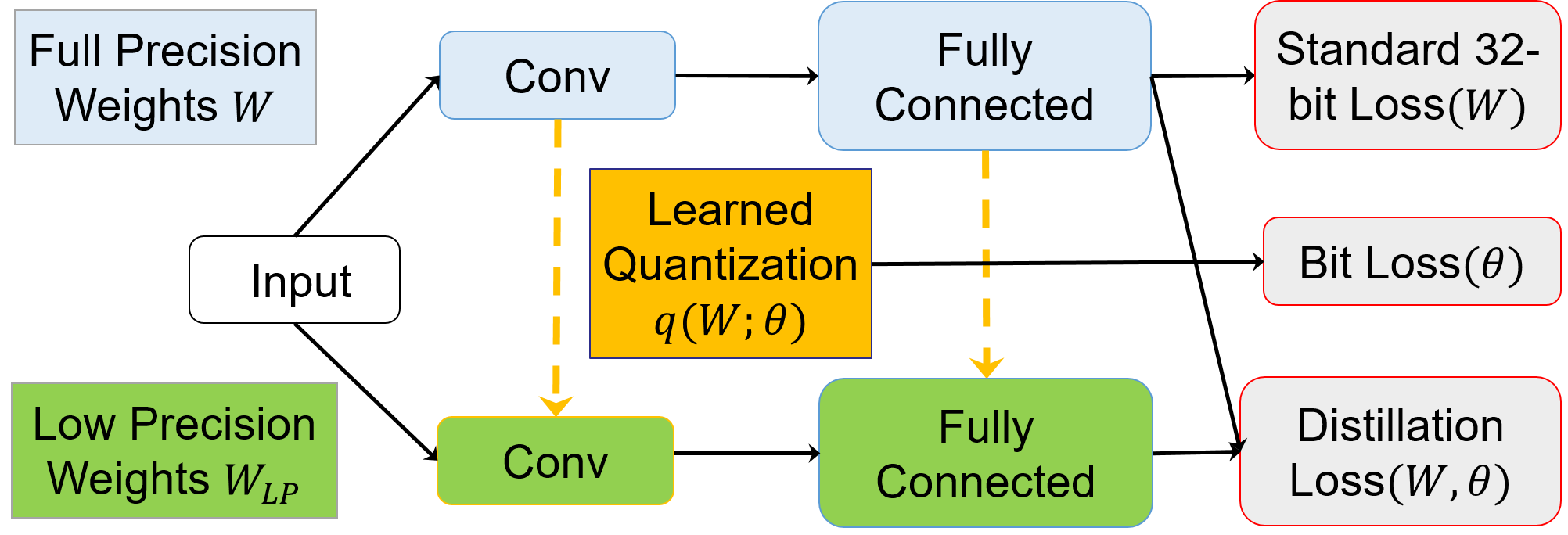}
    \caption{Illustration of the training scheme using $W$ to denote standard 32-bit floating point weights and $W_{LP}$ to denote the weights quantized according to (\ref{eq:quant}). The dashed arrows in the figure indicate that the values of $\theta_1, \theta_2$ as in (\ref{eq:quant}) are learned for each layer. The  novelty is that the number of bits and quantization levels are learned.}
    \label{fig:gtc_kd}
\end{figure}

We use Knowledge Distillation \cite{hinton2015distilling} motivated by the recent success \cite{polino2018model,mishra2018wrpn,mishra2017apprentice} of knowledge distillation techniques in training low-precision networks. Similar to the above prior work, we simultaneously train a standard 32-bit network, herein referred to as the teacher, and a low precision network, herein referred to as the student (see Figure \ref{fig:gtc_kd}). 

GTC differs from this prior work in three ways. First, GTC learns the number of bits as well as the quantized values. Second, while in previous work one goal of distillation is to reduce the number of neural network layers by using a compact student network distilled from a larger teacher network, note that GTC uses identical network architectures for the teacher and the student. Finally, in prior work the teacher is trained first and then fixed during the distillation process, whereas GTC learns both student and teacher models simultaneously from scratch. 
The GTC optimization objective is
\begin{equation}
\min_{\mathbf{W}, \mathbf{\theta}} \mathcal{L}(\mathbf{W}) + \lambda_1\mathcal{D}(\mathbf{W}, \theta) + \lambda_2\mathcal{B}(\theta)   \label{eq:loss} 
\end{equation}
The first term $\mathcal{L}$ is any standard supervised or unsupervised loss function, for example, cross-entropy (or negative log-likelihood) for supervised classification, or reconstruction error for unsupervised auto-encoders.

The second term $\mathcal{D}$ denotes a distillation loss that is a function of the pairwise predictions of the teacher and the student. Note that in our case, since the student parameters are derived from the teacher parameters and the quantization parameters in (\ref{eq:quant}), the distillation term is a function of $\mathbf{W}$ and $\mathbf{\theta}$. The difference of logits (before softmax) between the student and the teacher was used in \cite{hinton2015distilling}. Empirically, we found that using the cross-entropy (after softmax) leads to significantly better accuracy. We extended this to the unsupervised setting by using a sigmoid squashing function then  calculating the cross-entropy. 

The last term $\mathcal{B}$ penalizes values of $\theta$ that lead to a large variation in the power-of-two weights after quantization. This notion is captured in the number of bits as shown in (\ref{eq:quant_bits}), so $\mathcal{B}(\theta)=\sum_{l \in \text{layers}} 2^{\text{bits}^{(l)}}$ is simply the sum over all the neural network layers. Roughly speaking, $\lambda_2$ corresponds to the cost of bits and can be used to encode hardware constraints. As $\lambda_2 \rightarrow 0$, GTC may use up to 32-bits, and as $\lambda_2 \rightarrow \infty$, GTC produces a binary network. This penalty term is reminiscent of Minimum Description Length \cite{grunwald2007minimum} formulations. The regularizing effect of penalizing the number of bits was demonstrated in \cite{raghavan2017bitnet}. Note that in the absence of the penalty term, the student and teacher networks can be made arbitrarily close by adjusting the $\theta$ parameters.

One can interpret this objective as the Lagrangian of a constrained optimization problem \cite{raghavan2017bitnet,carreira2017model1,carreira2017model2}, that primarily optimizes the accuracy of the standard 32-bit model, with the constraints that the pairwise student-teacher predictions are similar and the student model is maximally compressed according as in Section \ref{sec:compression}. 

During training, we use the gradients wrt $\mathbf{W}$ and $\mathbf{\theta}$ to update\footnote{We use the ADAM optimizer unless otherwise specified.} $\mathbf{W}$ and $\mathbf{\theta}$ respectively, which implicitly determine the (quantized) parameters $\mathbf{\tilde{W}}$ of the student. Note that in general the gradients wrt $\mathbf{W}$ are not the same as the corresponding gradients in a standard 32-bit training due to additional gradients from the distillation loss. During testing, we dispose the teacher model $\mathbf{W}$ and make predictions using $\mathbf{\tilde{W}}$ alone. We use the identity function as the STE to propagate gradients through round and ceiling functions. 

It should be clear from the discussion that the above training procedure is conducted on a powerful GPU, and that the benefits of the power-of-two weights are not leveraged during training. That is, the training is performed offline before deployment to an edge device, where the compute, memory and power savings can be realized. Future work should extend GTC to quantization of gradients and activations e.g.\! using quantized backpropagation \cite{lin2015neural} or using projection-based algorithms. These are yet to be explored in the setting where the bits and levels are learned. 

\section{Experiments}
\label{sec:experiments}

We evaluate GTC on two tasks, namely,  supervised image classification and unsupervised image generation with a variational auto-encoder (VAE). We show results on two computer vision datasets: MNIST \cite{lecun1998gradient} and CIFAR-10 \cite{krizhevsky2009learning}. We used the ADAM optimizer \cite{kingma2014adam} with a learning rate denoted as $\mu$ unless otherwise specified. 

\textbf{Neural Architectures}: For the MNIST classification problem, we used a LeNet-style architecture that consists of four layers, namely two convolutional layers ($16$ and $36$ filters, each of size $5 X 5$, with $2 X 2$ max pooling and ReLU) followed by two fully connected layers ($128$ and $10$ outputs with ReLU and soft-max respectively). Note that this is a simpler model than LeNet-5. For CIFAR-10 classification, we used Configuration D of VGG-16 \cite{simonyan2014very}, with the addition of batch normalization and ReLU after convolutional layers. We also added Dropout \cite{srivastava2014dropout} after alternating convolutional layers to avoid overfitting. 

For the MNIST unsupervised problem, we only use the $28 \times 28$ images as input. The encoder part of the VAE uses three fully connected layers consisting of $512, 384$ and $256$ neurons respectively, each with tanh activations. The result is input in to two layers corresponding to the mean and variance, each with a latent dimension of $10$. The decoder consists of three fully connected layers consisting of $256, 384$ and $512$ neurons, each with tanh activations. Finally, we use one fully connected layer with $28 \times 28$ neurons with sigmoid activations to arrive at a $28 \times 28$ grayscale image. 

\subsection{Learning}
The first set of experiments demonstrate the training and convergence of GTC and the performance of the GTC model. In these experiments, we compare with the corresponding standard 32-bit model that is trained independently of GTC (ie.\! not the teacher model used in the distillation scheme of Section \ref{sec:learning}). In Section \ref{sec:prior}, we compare GTC with prior work as well as some natural baseline algorithms. 

\begin{figure*}[t]
    \centering
    \hfill
    \begin{minipage}{0.48\textwidth}
    \centering
    $\mu = 10^{-4}, \lambda_1 = 0.8, \lambda_2 = 0.04$.
    \includegraphics[width=0.9\textwidth, keepaspectratio]{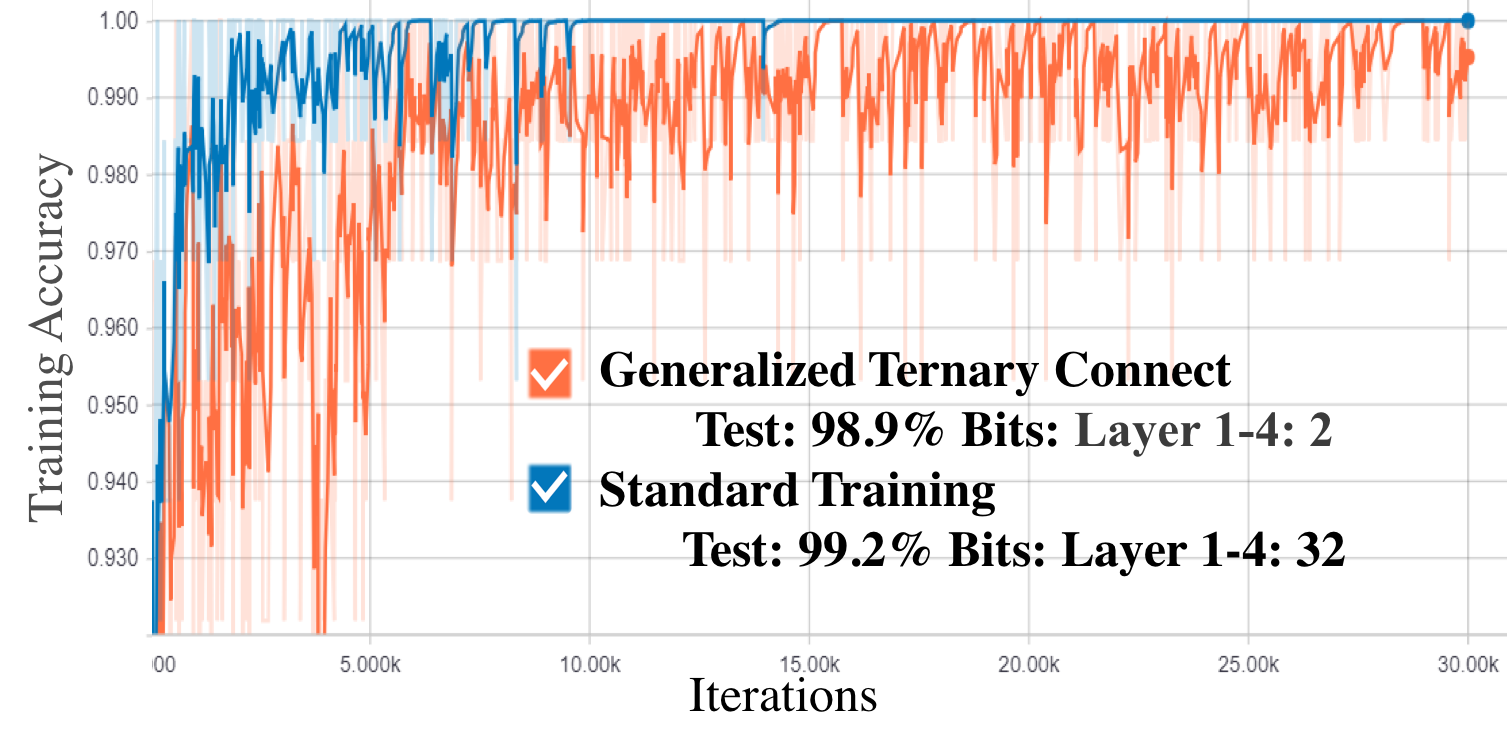}
    LeNet Network on the MNIST Dataset.
    \end{minipage}
    \hfill
    \begin{minipage}{0.48\textwidth}
    \centering
    $\mu = 10^{-5}, \lambda_1 = 0.8, \lambda_2 = 0.04$.    
    \includegraphics[width=0.9\textwidth, keepaspectratio]{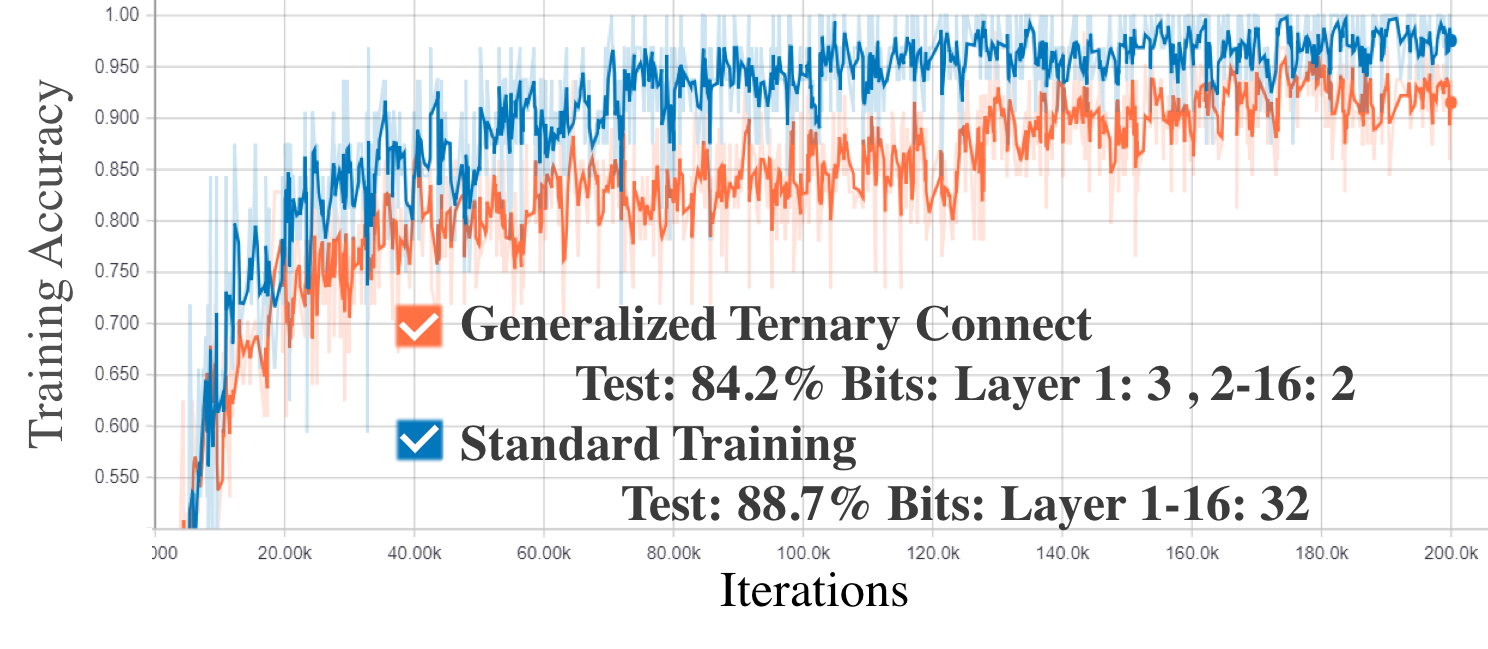}
    VGG-16 Network on the CIFAR-10 Dataset.
    \end{minipage}
    \hfill
    \caption{Comparison of accuracy over training iterations of GTC and the corresponding standard (32-bit) deep neural network.}
    \label{fig:eval_supervised}
    \hfill
    \begin{minipage}{0.44\textwidth}
    \centering
    $\mu = 10^{-3}, \lambda_1 = 3.0, \lambda_2 = 0.04$.
    \includegraphics[width=\textwidth, keepaspectratio]{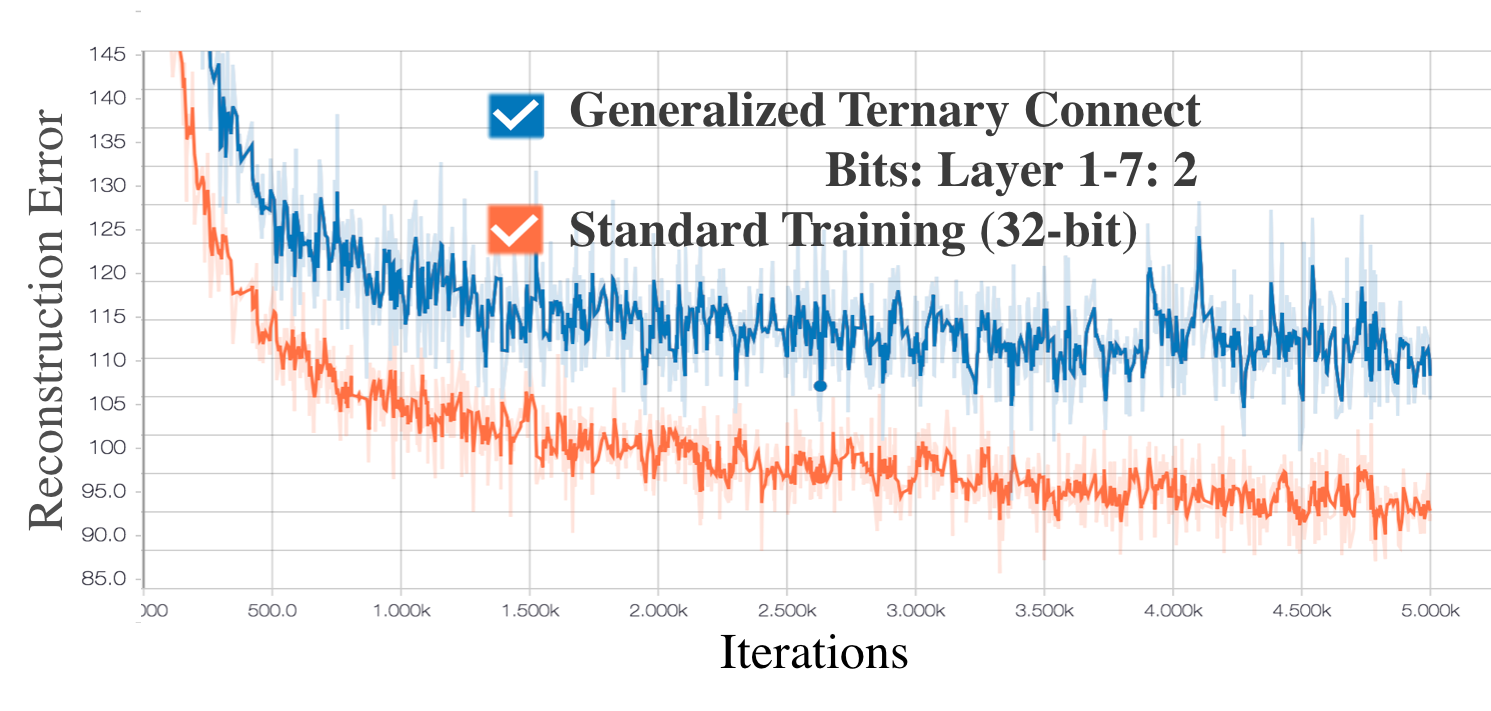}
    \caption{Comparison of the reconstruction error over iterations of GTC and the corresponding standard (32-bit) Variational Auto-Encoder on MNIST.}
    \label{fig:mnist_vae}
    \end{minipage}
    \hfill
    \begin{minipage}{0.48\textwidth}
    \centering
    SGD$(\mu=10^{-5})$, and GTC$(\lambda_1=0.8,\lambda_2=0.04)$, $\lambda_2$ divided by $10$ every $10K$ iterations.\\
    \includegraphics[width=\textwidth,keepaspectratio]{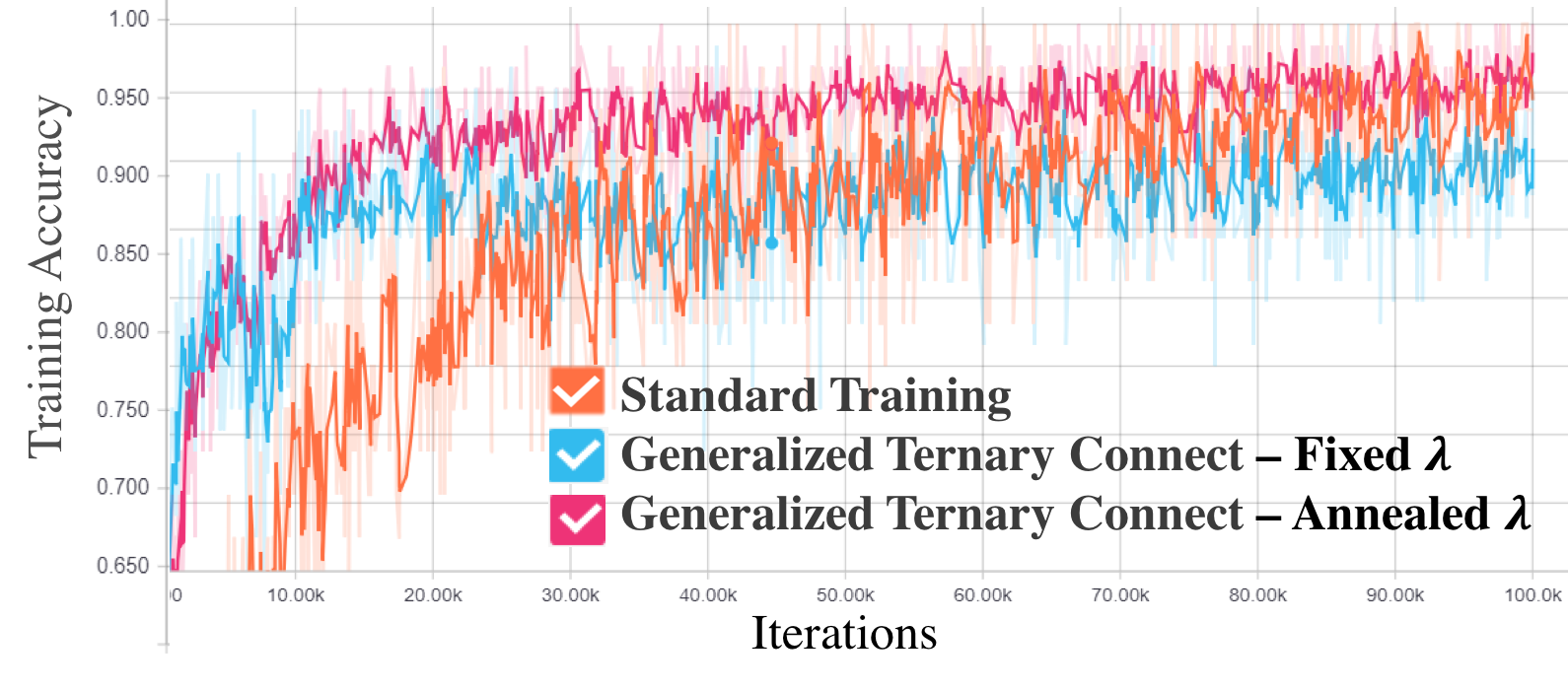}
    \caption{Experiment showing the use of GTC as an optimizer leads to faster convergence over Stochastic Gradient Descent (SGD) with a fixed learning rate. MNIST + LeNet.}
    \label{fig:mnist_anneal}
    \end{minipage}
    \hfill
    \begin{minipage}{0.25\textwidth}
    \centering
    \includegraphics[width=\textwidth,keepaspectratio]{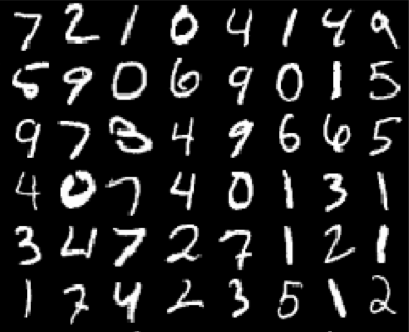}
    Groundtruth MNIST
    \end{minipage}
    \hfill
    \begin{minipage}{0.25\textwidth}
    \centering
      \includegraphics[width=\textwidth, keepaspectratio]{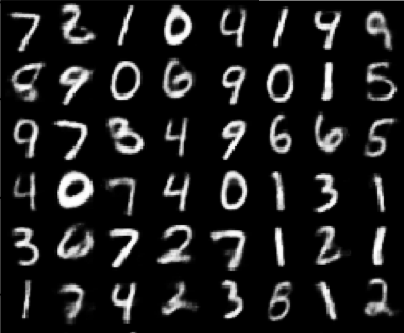}
    Reconstruction 32-Bit
    \end{minipage}
    \hfill
    \begin{minipage}{0.25\textwidth}
    \centering
    \includegraphics[width=\textwidth, keepaspectratio]{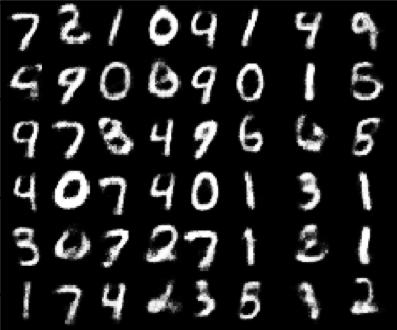}
    Reconstruction GTC
    \end{minipage}
    \hfill
    \caption{Visual inspection of the reconstruction of MNIST digits by the standard 32-Bit Variational Auto-Encoder and the GTC Variational Auto-Encoder.}
    \label{fig:reconstruction_mnist}
    \hfill
    \begin{minipage}{0.31\textwidth}
    \centering
    $\mu =10^{-4}, \lambda_1 =0.8, \lambda_2 = 0.04$.
    \includegraphics[width=\textwidth, keepaspectratio]{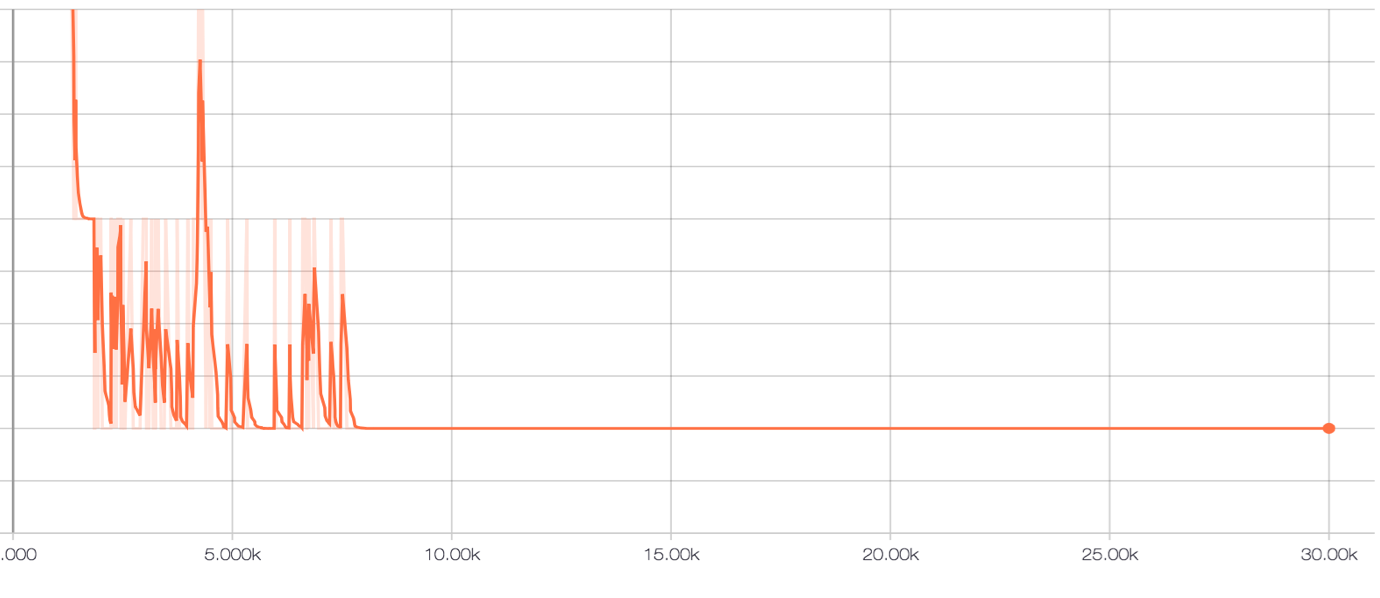}
    LeNet on MNIST.
    \end{minipage}
    \hfill
    \begin{minipage}{0.31\textwidth}
    \centering
    $\mu =10^{-5}, \lambda_1 =0.8, \lambda_2 = 0.04$.    
    \includegraphics[width=\textwidth, keepaspectratio]{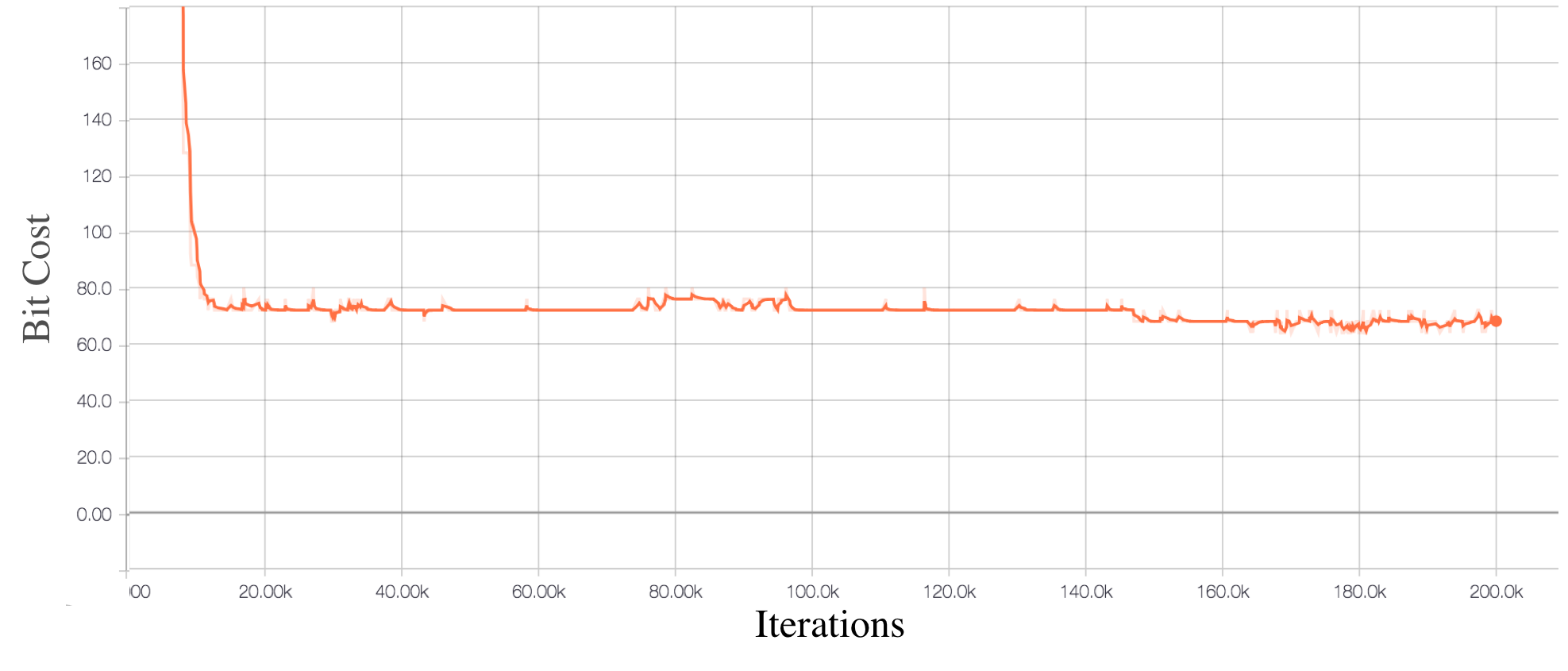}
    VGG-16 on CIFAR-10.
    \end{minipage}
    \hfill
    \begin{minipage}{0.31\textwidth}
    \centering
    $\mu = 10^{-3}, \lambda_1 = 3.0, \lambda_2 = 0.04$.
    \includegraphics[width=\textwidth, keepaspectratio]{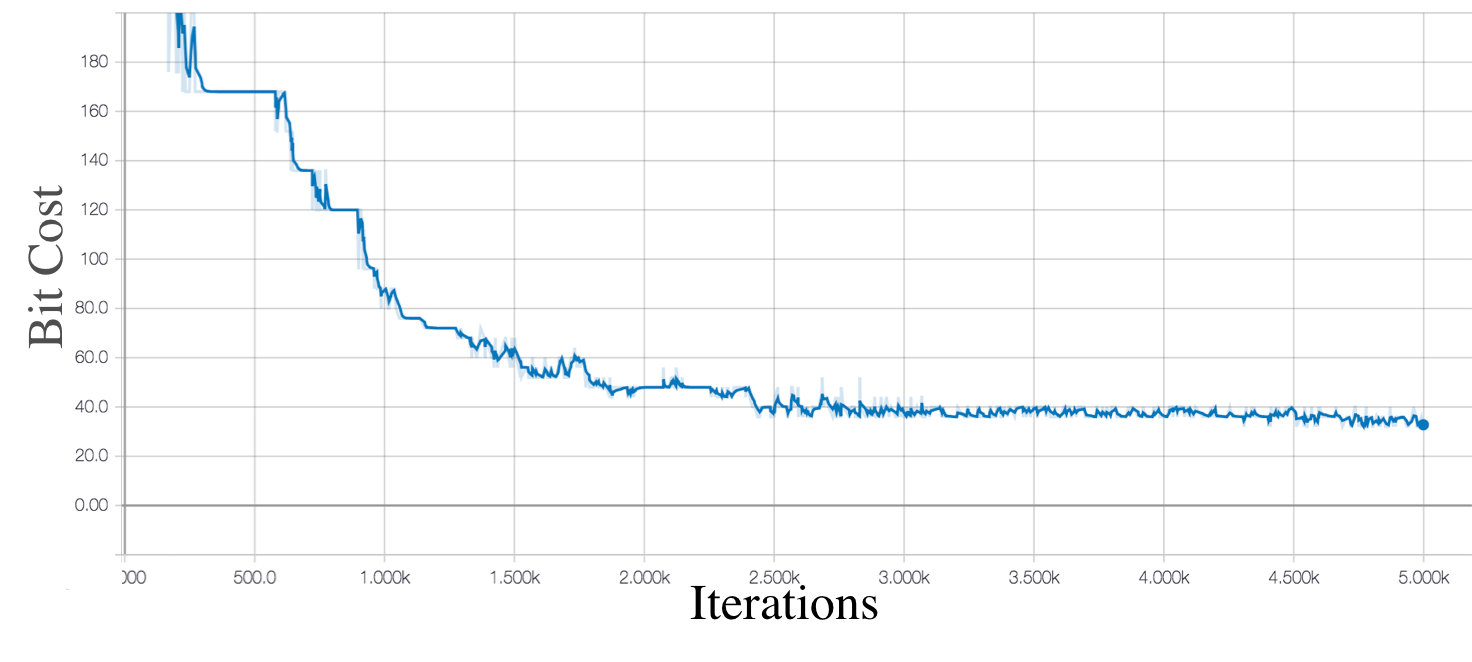}
    VAE on MNIST Dataset.
    \end{minipage}
    \hfill
    \caption{Convergence of the bit cost $\mathcal{B}(\mathbf{\theta})$ as in (\ref{eq:loss}), over training iterations of GTC on the MNIST and CIFAR-10 datasets.}
    \label{fig:eval_vae_bitcost}
\end{figure*}

Figure \ref{fig:eval_supervised} and Figure \ref{fig:mnist_vae} show these experiments for supervised classification and unsupervised VAE  respectively. 
In Figure \ref{fig:eval_supervised} (left) we see that GTC converges to a training accuracy within $1\%$ of the standard 32-bit model in roughly the same number of iterations (10K iterations). We note some oscillations in the training accuracy of GTC, however they are  within $1-2\%$ of the standard model once the standard model converges. This shows the benefit of training using distillation because the 32-bit model guides the learning of the GTC model. Moreover, on the test set, the accuracy of GTC is almost identical to that of the standard model. 

On the CIFAR-10 dataset (see Figure \ref{fig:eval_supervised} (right)), we observed a similar pattern. The training accuracy of the standard model was over $95\%$ whereas GTC achieved a training accuracy around $92\%$. The performance on the test set is lower than state-of-the-art results using VGG-16 on CIFAR-10.
We believe the lower performance on CIFAR-10 is caused  by  the  use  of  dropout  layers that tends to slow learning, limited  iterations of training, and no fine-tuning of the hyperparameters of GTC (see Section \ref{sec:hyper}). One can see that the trend in accuracy is upward. 

Figure \ref{fig:mnist_vae} shows a comparison of a VAE trained using GTC compared to a standard VAE for the MNIST dataset. For both of these variants, the reconstruction error was defined as the binary cross-entropy between the input image and the 32-bit reconstruction. For GTC, this is the first term as in (\ref{eq:loss}). The distillation loss (second term) was defined as the binary cross entropy between the reconstructions of the teacher and student models. The Figure shows that the VAE trained using GTC converges in terms of the reconstruction error. However, in some cases the visual quality of the reconstructions from GTC appear to be less blurry than the reconstructions of the 32-bit standard model (see Figure \ref{fig:reconstruction_mnist}). 

Overall, GTC naturally converged to binary networks for MNIST, and binary for all but the first layer of VGG-16 for the CIFAR-10 dataset, with very little loss of accuracy. GTC converged to a binary VAE network whose output has similar visual quality compared to the standard VAE. This is the main takeaway from the learning experiments on GTC. To the best of our knowledge, these are the first experiments demonstrating generative models such as VAE using low precision parameters. Future work may extend this to more sophisticated generative models and draw potentially different insights than feedforward classification models.
\subsection{Compression}
\begin{table}
\small
  \centering
    \caption{Weights learned by GTC shown as the number of bit shifts. A $\rightarrow$ ($\leftarrow$) sign indicates bits are shifted to the right (left). A $\lnot$ means the sign bit is flipped after shifting. For example, an entry $\lnot 0$ corresponds a weight of $-1$. A weight of zero is represented as $\emptyset$. Learned values of $\theta$ are shown.}
    \label{tab:weights_hist}
    $\begin{array}{|r|r|r|r|r|}
      \hline
      \multicolumn{5}{|c|}{\text{MNIST + LeNet}}\\
      \hline
      \textbf{Layer} & \textbf{Bits} & \text{Bit Shifts} & \theta_1 & \theta_2\\
      \hline
      \text{conv1} & 3 & [\rightarrow 4, \lnot \rightarrow 4, \rightarrow 5, \lnot \rightarrow 5] & -1.99& 0.05\\
      \hline
      \text{conv2} & 2 & [\rightarrow 2, \lnot \rightarrow 2] & -2.78& 0.04 \\
      \hline
      \text{fc1} & 2 & [\rightarrow 2, \lnot \rightarrow 2] & -2.70 & 0.04 \\
      \hline
      \text{fc2} & 2 & [\rightarrow 2, \lnot \rightarrow 2] & -2.70& 0.04\\
      \hline
      \multicolumn{5}{|c|}{\text{CIFAR-10 + VGG-16}}\\
      \hline
      \text{conv1} & 3 & [\rightarrow 1, \rightarrow 2, \lnot \rightarrow 2, \rightarrow 3, \lnot \rightarrow 3] & 2.45& 0.03\\
      \hline
      \text{conv2} & 2 & [\rightarrow 2, \lnot \rightarrow 2, \rightarrow 3, \lnot \rightarrow 3] &  -1.11& 0.19\\
      \hline
      \text{conv3} & 2 & [\rightarrow 0, \lnot \rightarrow 0, \rightarrow 1, \lnot \rightarrow 1] & -0.13 & 0.05\\
      \hline
      \text{conv4} & 2 & [\rightarrow 1, \lnot \rightarrow, \rightarrow 2, \lnot \rightarrow 2] & -0.37&  0.16\\
      \hline
      \text{conv5} & 2 & [\rightarrow 0, \lnot \rightarrow 0]  & 0.01& 0.03 \\
      \hline
      \text{conv6} & 2 & [\rightarrow 0, \lnot \rightarrow 0] & -0.07& 0.02 \\
      \hline
      \multicolumn{5}{|c|}{\ldots} \\
     \hline
     \text{fc1} & 2 & [\rightarrow 0, \lnot \rightarrow 0] & -0.09& 0.01 \\
     \hline
     \text{fc2} & 2 & [\rightarrow 1, \lnot \rightarrow 1, \emptyset] & 0.03 & 0.08\\
     \hline
     \text{fc3}  & 2 & [\rightarrow 6, \lnot \rightarrow 6, \rightarrow 5, \lnot \rightarrow 5] & -0.93& 0.85\\
     \hline
      \multicolumn{5}{|c|}{\text{MNIST + VAE}}\\
      \hline
      \text{fc1} & 2 & [\rightarrow 0, \lnot \rightarrow 0, \rightarrow 1, \lnot \rightarrow 1] & 0.15& 0.06\\
      \hline
      \text{fc2} & 2 & [\rightarrow 0, \lnot \rightarrow 0, \rightarrow 1, \lnot \rightarrow 1] & -0.03 & 0.09\\
      \hline
      \text{fc3} & 2 & [\rightarrow 1, \lnot \rightarrow  1, \rightarrow 2, \lnot \rightarrow 2] & -0.01 & 0.33 \\
      \hline
       \text{fc4} & 2 & [\rightarrow 2, \lnot \rightarrow  2, \rightarrow 3, \lnot \rightarrow 3] & -0.14 &  0.28\\
       \hline
       \multicolumn{5}{|c|}{\ldots} \\
       \hline
\end{array}$
\end{table}
In order to further showcase the compression achieved by GTC, Table \ref{tab:weights_hist} shows the weights learned by GTC along with the number of bits and the learned values of $\theta_1$ and $\theta_2$. First, we observe that the learned $\theta_1$ values tend to be negative, whereas the $\theta_2$ values tend to be positive. That is, GTC learns S-shaped quantization functions. A special case of this is seen in the middle layers of VGG-16, where both $\theta$ values that are close to zero, which corresponds to a Heaviside function (see Figure \ref{fig:quant_example}). 

The learned weights are shown as the number of bit shifts (and flipping of sign bit) required so that the result is equivalent to a floating-point multiplication. For each number of shifts in each layer, GTC uses both the signed (with $\lnot$ flip) and unsigned weight. The histograms show that the learned weights are almost always bi-modal with two equal modes, and never uni-modal. The weights of the middle layers of VGG are $\pm 1$ shown as zero bit shifts. 

Figure \ref{fig:eval_vae_bitcost} shows the convergence of the bit cost $\mathcal{B}(\mathbf{\theta})$ as in (\ref{eq:loss}). The figure shows the sum over layers of the number $2^\text{bits}$ of bits used. We observe some oscillations for LeNet on MNIST, but no oscillations are seen in VGG-16 on CIFAR-10 as well as VAE on MNIST. In all three cases, the bit cost converges within the first $30\%$ of the total training iterations. 
\subsection{Experimental Comparison to Prior Work}
\label{sec:prior}
\begin{table}
\small
\centering
\caption{Table comparing the accuracy and compression of GTC to prior work. Superscript ${}^*$ denotes numbers taken from the corresponding paper. Subscript ${}_{\#}$ denotes a neural architecture that is slightly different than GTC. A hypen ($-$) indicates that the paper used a completely different architecture or did not evaluate on these datasets.}
$\begin{array}{|c|c|c|c|c|} 
 \hline
 & \multicolumn{2}{|c|}{\text{LeNet/MNIST}} & \multicolumn{2}{|c|}{\text{VGG-16/CIFAR-10}} \\
\text{Method} & \text{Acc. (\%)} & \text{Avg. Bits} & \text{Acc. (\%)} & \text{Avg. Bits} \\
\hline
\text{PM} & 99.2 & 9 & 80.1 & 9\\
\text{STE} &  99.0 & 9 & 88.4 & 9\\
\text{STE+Bit} & 98.4 & 2 & 44.7 & 2\\
\text{LogNet} & 10.3 & 5 & 17.1 & 5\\
\text{LogNet}^* & - & - & 93^* & 5^* \\
\textbf{GTC} & \textbf{98.9} & \textbf{2} & \textbf{84.2} & \textbf{2} \\
\text{TC}_\# & 98.85^* & 2^* & - & -\\
\text{DC}_\# & 99.2^* & 5.3^* & - & - \\
\text{BC}_\# & 98.2^* & 11.75^* & \textbf{91.8}^* & 9.2^*\\
\text{VQ}_\# & \textbf{99.2}^* & \textbf{2}^* & - & -\\
\text{DQ}_\# & - & - & 80^* & 2^*\\
\text{BitNet}_\# & 97^* & 6^* & - & -\\
\hline
\end{array}$
\end{table}

We compared GTC with the following.
\begin{itemize}
    \item Post Mortem (PM) takes the final 32-bit model and snaps each weight to the nearest integer power of two. The result requires $9$ bits of storage ($8$ bits for exponent, $1$ bit for sign). In general, there is no guarantee of accuracy.
    \item Straight Through Estimator (STE) directly trains the quantized model without using distillation. The objective function is the cross-entropy of the student network. Gradients wrt $\mathbf{W}$ are calculated using identity STE and used to update $\mathbf{W}$. There are no $\theta$ parameters. This approach worked very well in terms of accuracy, but requires $9$ bits. 
    \item STE+Bit uses the same approach as above, with the addition of a bit penalty term $\mathcal{B}$ as in (\ref{eq:loss}). The bits are calculated using (\ref{eq:quant_bits}), without using any $\theta$ parameters. This approach retained the high accuracy of STE in MNIST while compressing to a binary network. It did not make any progress on the CIFAR-10 dataset and resulted in poor accuracy. Together, these indicate the necessity of the distillation approach to training. 
    \item LogNet \cite{miyashita2016convolutional,lee2017lognet} proposed a method to train a network similar to GTC, but weights, gradients and activations of the network are quantized. The number of bits is not learned. We attempted to reproduce their algorithm but restricted the quantization to the weights of the network. We were not able to reproduce the performance reported. 
    \item Ternary Connect TC \cite{lin2015neural}: We show the results from the paper. Their network architecture is slightly different from the one used for the GTC experiments. GTC is more general and shows superior compression to TC.
    \item Bayesian Compression (BC) \cite{louizos2017bayesian}: We show the performance as reported. BC shows lower accuracy on MNIST and higher accuracy on CIFAR-10. We note that the VGG-16 implementation therein does not use any dropout. GTC shows significantly higher compression than BC.
    \item Variational Quantization (VQ) \cite{achterhold2018variational} extends BC to ternary quantization. The number of bits is not learned. VQ shows similar accuracy and compression as GTC on MNIST. 
    \item Deep Compression (DC) \cite{han2015deep} does not learn the number of bits, but learns the locations of the centroids. It uses quantization error as the primary criterion. GTC shows superior accuracy and compression.
    \item Differentiable Quantization (DQ) \cite{polino2018model} does not learn the number of bits, but learns the locations of the centroids. The paper used a ResNet-style architecture on CIFAR-10 with a fixed $2$ bits. The reported $80\%$ accuracy is inferior to the accuracy of GTC for the bits. 
    \item BitNet \cite{raghavan2017bitnet} does learn the number the bits on a linear quantization scale, and has a similar learning scheme as GTC. However, the reported accuracy is lower with lower compression in comparison to GTC.
    \item GTC (this work): Over all the compared methods, GTC shows the best compression with minimal drop in accuracy. We believe the lower performance on CIFAR-10 is caused by the use of dropout layers, limited iterations of training, and no fine-tuning of the  hyperparameters of GTC. Most of the above baselines, with the exception of PM, STE and TC do not result in a neural network whose forward pass does not require any multiplications. 
\end{itemize}
\subsection{Hyperparameters}
\label{sec:hyper}
\begin{table}    
\small
\centering
\caption{Accuracy and Average Number of Bits vs Hyperparameters on MNIST+LeNet: rows show $\lambda_2$ (bit penalty) and columns show $\lambda_1$ (distillation penalty). Trends in the table show the trade-off of precision vs accuracy in GTC.}
$\begin{array}{|c|c|c|c|c|c|c|} 
 \hline
 \lambda_2 / \lambda_1 & 0.0 & 0.2 & 0.4 & 0.6 & 0.8 & 1.0\\ 
 \hline
 \multicolumn{7}{|c|}{\text{Average Number of Bits}}\\
 \hline
 0.0 & 5.75 & 5.25 & 5.25 & 5.25 & 5.25 & 5.25\\ 
10^{-7} & 2 & 4 & 4.25 & 4 & 4.25 & 4.25 \\ 
10^{-6} & 2 & 3.75 & 3.75 & 4 & 4 & 4 \\ 
10^{-5} & 2 & 2.75 & 3 & 3 & 3.25 & 3.5 \\
10^{-4} & 2 & 2.5 & 2.5 & 2.25 & 2.25 & 2.5 \\
10^{-3} & 2 & 2 & 2 & 2 & 2 & 2 \\ 
 \hline
 \multicolumn{7}{|c|}{\text{Accuracy}}\\
 \hline
 0.0 & 99.3 & 99.3 & 99.3 & 99.3 & 99.3 & 99.3\\ 
 10^{-7} & 96.9 & 99.3 & 99.3 & 99.2 & 99.3 & 99.3 \\ 
 10^{-6} & 94.5 & 99.2 & 99.3 & 99.2 & 99.2 & 99.2 \\ 
10^{-5} & 94.2 & 99.1 & 99.2 & 99.1 & 99.2 & 99.1 \\
10^{-4} & 96.7 & 98.7 & 99.0 & 99.0 & 99.0 & 99.0 \\ 
10^{-3} & 95.0 & 98.4 & 98.5 & 98.7 & 98.8 & 98.6 \\ 
 \hline
\end{array}$
\label{tab:hyper}
\end{table}
In previous sections, we showed the results of GTC with a fixed value of $\mu, \lambda_1$ and $\lambda_2$. We arrived at these values by using a grid search on the MNIST dataset and picking the setting with highest accuracy with lowest number of bits. We did not perform any tuning for the CIFAR-10 dataset, which might be the reason for the somewhat low performance. Table \ref{tab:hyper} shows the results of this grid search. As expected, for any fixed value of $\lambda_1$, as the value of $\lambda_2$ increases, the number of bits quickly decreases to $2$ and the accuracy decreases significantly. Similarly, for any fixed value of $\lambda_2$, increasing the coefficient for distillation loss $\lambda_1$ increases the number of bits, and generally increases the accuracy. 
Ultimately, hardware constraints dictate the desired number of bits, which in turn can be used to decide appropriate values for $\lambda_1$ and $\lambda_2$. Next we show an interesting use case of GTC as an optimizer by annealing the values of $\lambda_2$.
\subsection{GTC for Optimization}
Prior work \cite{raghavan2017bitnet,lin2015neural} has noted that low precision networks exhibit fast convergence in the early stages of learning in some cases. We leverage this property in a novel variant of GTC that can be used as an off-the-shelf optimizer. We initialize GTC with a high value of $\lambda_2$, the coefficient of bit cost and as training progress, we rapidly reduce the value of $\lambda_2$.  
Figure \ref{fig:mnist_anneal} shows the result of such a GTC variant. We used Stochastic Gradient Descent (SGD) with a fixed learning rate $\mu$. In the variant denoted as Annealed $\lambda$, $\lambda_2$ is divided by $10$ every $10000$ iterations. The figure shows that the training accuracy of both of the variants of GTC is higher till 40K iterations. The 32-bit standard network (optimized by SGD) converges to a higher accuracy over GTC with a the fixed $\lambda_2$ after 40K iterations. Interestingly, the variant with annealed $\lambda_2$ has superior accuracy and dominates the 32-bit standard model throughout. It is to be noted that this effect vanishes when using ADAM instead of SGD, confirming the effect of low precision training on the effective learning rate as noted in \cite{raghavan2017bitnet}.


\section{Impact}
\label{sec:discussion}
This paper proposed GTC that generalizes ternary connect in a number of ways. GTC was demonstrated to be effective in learning and compression, and in comparison to baselines and prior work. These algorithmic advances directly translates to power and computational efficiency during inference. To quantify the gains in hardware, we offer two key simulations on FPGA and Raspberry-Pi. These gains comes from the compressed weights and the elimination of multiplications. Instead of performing the standard multiply/add operation, we can instead do a shift/add operation. On FPGA, the per operation power consumption reduced from 49mW for multiplication to 10mW for bit-shifting as in (\ref{eq:mult_shift_add})\footnote{Based on RTL synthesis at 800MHz on 16nm Zynq Ultrascale+(ZU9EG) using Xilinx Vivado Tools}. Similarly on Raspberry-Pi, the per operation time taken reduced from 93ns to 67ns. These gains are significant when amortized over the millions or billions of parameters in deep networks. Additional gains are anticipated from handling, storing and retrieving compressed weights. Specialized digital hardware can implement bit-shifting directly with memory storage for further gains. Even in analog multipliers, a reduction in circuitry is realized with simpler current/voltage dividers compared to complex op-amp circuitry. 

\clearpage
\section{Acknowledgements}
This material is based upon work supported by the Office of Naval Research (ONR) under contract N00014-17-C-1011, and NSF \#1526399. The opinions, findings and conclusions or recommendations expressed in this material are those of the author and should not necessarily reflect the views of the Office of Naval Research, the Department of Defense or the U.S. Government.
\bibliographystyle{aaai}
\fontsize{9.0pt}{10.0pt} 
\selectfont
\bibliography{references}
\end{document}